\documentclass{article}
\usepackage{ijcai16}
\usepackage[dvipsnames]{xcolor}
\usepackage{times}
\usepackage{url}
\usepackage[implicit=false]{hyperref}

\usepackage[lofdepth,lotdepth]{subfig}
\usepackage{graphicx}
\usepackage{xspace}
\newcommand{\cut}[1]{}
\newcommand{\namecite}[1]{\citeauthor{#1} [\citeyear{#1}]}
\newcommand{\gamerelated}{game-related\xspace}

\title{Tie-breaker: Using language models to quantify gender bias in sports journalism}
\author{Liye Fu \and Cristian Danescu-Niculescu-Mizil \and  Lillian Lee \\
Cornell University \\
 {liye@cs.cornell.edu \hspace*{.18in} cristian@cs.cornell.edu \hspace*{.18in} llee@cs.cornell.edu}}

\begin{document}

\maketitle

\begin{abstract}

Gender bias is an increasingly important issue in sports journalism.
In this work, we propose a language-model-based approach to quantify differences in questions posed to
female
vs.~male athletes, and apply it to tennis post-match interviews.
 We find that journalists ask male players questions that are generally more focused on the game when compared with the questions they ask their female counterparts. We also provide a fine-grained analysis of the extent to which the salience of this bias depends on various
 factors, such as question type, game outcome or player rank.

\end{abstract}

\section{Introduction}

There has been an increasing level of attention to and discussion of gender bias in sports,  ranging from differences in pay and prize money\footnote{``U.S. Women, Fighting for Equal Pay, Win Easily as Fans Show Support'', \textit{The New York Times},  April 6, 2016. \scriptsize{\url{http://www.nytimes.com/2016/04/07/sports/soccer/uswnt-colombia-friendly-equal-pay-complaint.html}}} to different levels of focus on off-court topics in interviews by journalists. With respect to the latter, Cover the Athlete,\footnote{\scriptsize{\url{http://covertheathlete.com/}}}
 an initiative that urges the media to focus on sport performance, suggests that female athletes tend to get more ``sexist commentary" and ``inappropriate interview questions"
than males do; the organization put out an attention-getting video in 2015 purportedly showing male athletes' awkward reactions to receiving questions like those asked of female athletes.
However, it is not universally acknowledged that female athletes attract more attention for off-court activities.
For instance, a
manual analysis by \namecite{Kian:JournalOfBroadcastingElectronicMedia:2009} of online articles revealed significantly more descriptors associated with the physical appearance and personal lives of male basketball players in comparison to female ones.

Transcripts of pre- or 
post-game
press conferences offer an opportunity to determine quantitatively
and in a data-driven manner
 how different are the questions  which journalists pose 
 to male players from those they pose to female players.
  Here are  examples of a 
  \gamerelated
   and a non-game-relevant
  question,  respectively, drawn from actual tennis interviews:

\begin{enumerate}
\item What happened in that fifth set, the first three games?
\item After practice, can you put tennis a little bit behind you and have dinner, shopping, have a little bit of fun?
\end{enumerate}

To quantify gender discrepancies in questions, we propose
a statistical language-model-based approach to measure how 
\gamerelated
questions are. 
In order
 to make such an approach effective, we restrict our attention in this study to a single sport---tennis---so that mere variations in the lingo of different sports do not introduce extra noise in our language models. Tennis is also useful for our investigation because, as \namecite{kian2011} noted, it ``marks the only professional sports where male and female athletes generally receive similar amounts of overall broadcast media coverage during the major tournaments."

Using our methodology, we are able to quantify
gender bias
with respect to how \gamerelated interview questions are. 
  We also provide a more fine-grained analysis %
  of how gender differences
in journalistic questioning
 are displayed under various scenarios. To help with further analysis %
 of interview questions and answers, we introduce a dataset of tennis post-match interview transcripts along with
corresponding
 match information.\footnote{Dataset available at \scriptsize{\url{http://www.cs.cornell.edu/~liye/tennis.html}}}

\section{Related Work}

In contrast with our work,
prior investigations of bias in sport journalism
rely on manual coding or are based on simple lists of manually defined keywords.
These focus on bias with respect to race, nationality, and gender
\cite{Rainville:JournalismMassCommunicationQuarterlyJournalism:1977,sabo96,Eastman:HowardJournalOfCommunication:2010,Bruce01112004,Billings:OlympicMediaInsideTheBiggestShow:2008,kian2011,livcen2013cheering};
see \namecite{van2010race} for a review.

Much of the work
on gender bias in sports reporting has focused on
``air-time''
\cite{eastman2000sportscasting,higgs2003gender}.
Other 
studies
 looked
at stereotypical descriptions and framing
\cite{Messner:GenderSociety:1993,jones2004half,angelini2010agenda,Kian:JournalOfBroadcastingElectronicMedia:2009}.
For surveys, see
\namecite{Knight:SexRoles} or \namecite{Kaskan:SexRoles:2014}, inter alia. 
Several studies have focused on the particular case of gender-correlated differences in tennis coverage
\cite{Hilliard:SociologyOfSportJournal:1984,Vincent:InternationalJournalOfSportManagementAndMarketing:2007,kian2011}. We extend this line of work by proposing an automatic way to quantify gender bias in sport journalism.

\section{Dataset Description}\label{sec:data}

We collect tennis press-conference transcripts from ASAP Sport's website
(\url{http://www.asapsports.com/}),
whose tennis collection dates back to 1992 and is still updated for current tournaments. For our study, we take post-
game
 interviews for tennis singles matches played between Jan, 2000 to Oct 18, 2015. We also obtain easily-extractable match information from a dataset provided by Tennis-Data,\footnote{
\scriptsize{\url{http://www.tennis-data.co.uk/}}}
 which
covers the majority of the
matches
played on the men's side from 2000-2015 and on the women's side from 2007-2015.

We match interview transcripts with 
game statistics by date and player name, keeping only the question and answer pairs from 
games
 where the statistics are successfully merged. This gives us a dataset consisting
of
 6467 interview transcripts and a total of 81906 question snippets\footnote{Each snippet represents one turn from one journalist. Most question snippets contain at least one question, although some could be merely clarifications or comments. Note that reporter information (who asked which question) is not available in the transcript.}
posed to 167 female players and 191 male players. %

To model
{\em tennis-game}-specific language, we use
live text
play-by-play
commentaries
collected from the website Sports Mole
(\url{http://www.sportsmole.co.uk/}).
These tend to be short, averaging around %
40 words. Here is a sample,
taken from
the
Federer-Murray match at
the
2015 Wimbledon semi-final:\footnote{
\scriptsize{\url{http://www.sportsmole.co.uk/tennis/wimbledon/live-commentary/live-commentary-roger-federer-vs-andy-murray-as-it-happened_232822.html}}}
\begin{quote}
``The serve-and-volley is being used frequently by Federer and it's enabling him to take control behind his own serve. Three game points are earned before an ace down the middle seal
[sic]
the love
hold.''
\end{quote}
For our analysis, we create a 
gender-balanced
 set of commentaries consisting of descriptions for 1981
 games played for each gender.

\section{Method}\label{sec:method}

As a preliminary step, we apply a word-level analysis to understand if there
appear to be
differences in word usage when journalists interview male players compared to female players. We then introduce our method for quantifying the degree to which a question is 
\gamerelated,
 which we will use to explore gender differences.
\subsection{Preliminary Analysis}
To compare word usage in questions, we consider, for each word $w$, the percentage of players who have ever been asked a question containing $w$.
We then consider words with the greatest difference 
in percentage between male and female players.\footnote{Words that are gender-specific 
(like `her')
are manually discarded.} The top distinguishing words,
which are listed below in descending order of percentage difference,
seem to suggest that questions journalists pose to male players are more 
\gamerelated:

\begin{description}
 \item[Male players: ] clay, challenger(s), tie, sets, practiced, tiebreaker, maybe, see, impression, serve, history, volley, chance, height, support, shots, server(s), greatest, way, tiebreaks, tiebreakers, era, lucky, luck;\footnote{
It is interesting, but beyond the scope of this paper, to speculate on reasons why ``luck'' and ``lucky'' skew so strongly male.
 }
 \item[Female players: ] yet, new, nervous, improve, seed, friends, nerves, mom, every, matter, become, meet, winning, type, won, draw, found, champion, stop, fight, wind, though, father, thing, love.
\end{description}
\subsection{Game Language Model}

To quantify how \gamerelated a question is
in a data-driven fashion,
we train a bigram language model using KenLM%
\footnote{KenLM ({\scriptsize \url{https://kheafield.com/code/kenlm/}}) 
estimates language models using modified Kneser-Ney smoothing without pruning.}
\cite{kenlm}
on the gender-balanced set of live-text
play-by-play
commentaries introduced in Section \ref{sec:data}. 

For an individual question
$q$, we measure its
{\em perplexity} $PP(q)$ with respect to this {\em 
game language model}
$P_{\textnormal{\tiny \tiny commentary}}$ as an indication of how 
\gamerelated
 the question is: the higher the perplexity value, the less \gamerelated the question.
Perplexity,
a standard measure of language-model fit \cite{jelinek1977perplexity},  is
defined
as follows
for an $N$-word sequence $w_1 w_2 \ldots w_N$:

\[
PP(w_1 w_2 ... w_N) = \sqrt[N]{\displaystyle\frac{1}{P_{\textnormal{\tiny \tiny commentary}}(w_1\cdots w_N)}} \hspace*{.1cm}.
\]
Below are some sample questions of low-perplexity and high-perplexity values:
\begin{center}
\begin{tabular}{  c | c  }
 \hline
 {\bf Perplexity} & {\bf Sample Questions}  \\
 \hline
Low      &  What about your serve, Rafa? \\
            	  & The tiebreak, was that the key to the match? \\
\hline
High 	& Who designed your clothes today? \\
  & Do you normally watch horror films to relax? \\
 \hline
\end{tabular}
\end{center}

\section{Experiments}

In this section we use the game language model to quantify gender-based bias in questions.
We then compare the extent to which this difference depends of various
 factors, such as question type,
 game outcome, or player rank.
\subsection{Main Result: Males vs.~Females}

We first compute perplexities for each individual question\footnote{We identify individual
questions simply by looking for `?'.} and then group the question instances according to the interviewee's gender class.
Throughout we use the Mann-Whitney $U$ statistical significance test,\footnote{
We used this non-parametric significance test instead of the $t$-test because it doesn't assume the samples to be normally distributed.}
unless otherwise noted.

Comparing perplexity values between the two groups, we %
 find that
{\em the mean perplexity of questions posed to male players is significantly smaller ($p$-value $<$0.001)
 than that of questions posed to female players.  This suggests that the questions male athletes receive are more \gamerelated.}

However, the number of interviews each player
participates
in varies greatly, with highly interviewed players answering as many as thousands of questions while some lesser-known players have fewer than 10 interview questions in the dataset.  Thus it is conceivable that the difference is simply explained by questions asked to a few prolific players.  To test whether this is the case, or whether the observation is more general, we micro-average the perplexities by player: for each of the 167 male players and 143 females who have at least 10 questions in our dataset, we consider the average perplexities of the questions they receive.
Comparing these micro-averages, we find that it is still the case that questions posed to male players are significantly closer to game language ($p$-value $<$ 0.05), 
{\em  indicating that 
the %
observed
 gender difference is not simply explained
by 
 a 
few
highly interviewed players.}

\subsection{Relation to Other Factors}

We further investigate how the level of gender bias is tied to different factors:
how
typical the question is
(section \ref{sec:typical}), the ranking of the player (section \ref{sec:ranking}), and
whether the player won or lost the match (section \ref{sec:winning}).
For all the following experiments, we use
per-question perplexity for
comparisons:
per-player perplexity is not used due to limited sample size. %

\subsubsection{Typical vs.~Atypical Questions}
\label{sec:typical}

One might wonder whether the perplexity disparities we see in questions asked of female vs.~male players are due to ``off-the-wall'' queries, rather than to those that are more typical in post-match interviews.
We
therefore
use a data-driven approach to distinguish between {\em typical} and {\em atypical} questions.

For any given question, we consider how frequently its words appear in post-match press conferences in general. Specifically, we take the set of all questions as the set of
documents, $D$. We compute the inverse document frequency for each word (after stemming) that has appeared in our dataset, excluding the set $S$ consisting of stop words and a special token for entity names.\footnote{We
replace capitalized words and phrases with ``$<$NOUN$>$'';
for
each word at the beginning of a sentence (which is always capitalized),
we check whether it is a dictionary word.} For a question $q$ that contains the set of unique words $\{w_1, w_2,  ... , w_N\}\notin S$, we compute its {\em atypicality} score $Sc(q)$
as:
\[
Sc(\{w_1, w_2,  ... , w_N\}) = \displaystyle\frac{1}{N}\sum\limits_{i=1}^{N} \textnormal{idf}(w_i, D) \, .
\]

We use the overall mean atypicality score of the entire question 
dataset
 as the cutoff point: questions with scores above the overall mean are considered atypical and the rest are considered
typical.\footnote{Questions consisting only
of
stop words and player or tournament names are still considered
typical questions, even though they do not have 
an
 atypicality score. } Below are some examples:

\begin{center}
\begin{tabular}{  c | c  }
 \hline
 {\bf Category} & {\bf Sample Questions}  \\
 \hline
Typical      &  Have you played each other before? \\
            	  & How do you feel playing here?\\
\hline
Atypical & What about your haircut?  \\
	& Are you a vodka drinker? \\
 \hline
\end{tabular}
\end{center}

\begin{figure}[h]
 \centering
   \includegraphics[width=0.4\textwidth]{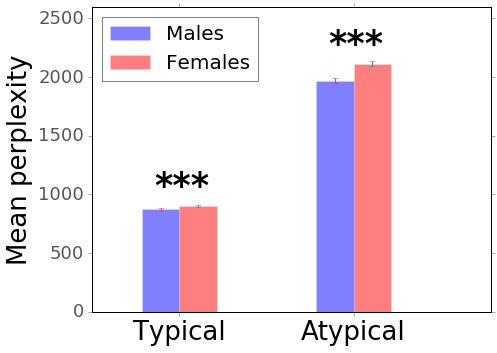}
   \label{fig:factorsa}
\caption{Mean perplexity values for
male and female athletes after grouping the questions by
how typical they are.
Stars indicate
high
statistical significance ($p<0.001$)
between the male and female case.
The male-female difference for the atypical group is statistically significantly larger than for the typical group.
}
\label{fig:factors-typical}
\end{figure}

Figure \ref{fig:factors-typical} shows
that a gender bias
with respect to whether \gamerelated language is used
 exists for both typical and atypical questions.
However, additional analysis reveals that
the difference in mean perplexity values between genders is
highly statistically significantly
larger for atypical questions, suggesting that gender bias is more salient among the more
unusual
queries.

\subsubsection{Player Ranking}
\label{sec:ranking}

\begin{figure}[h]
\centering
   \includegraphics[width=0.4\textwidth]{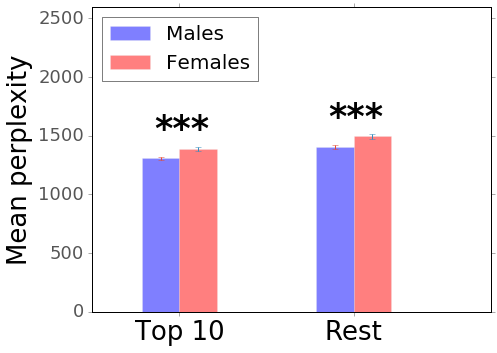}
\caption{Mean perplexity values for male and female athletes after grouping the questions by
the ranking of the player to which they are addressed. Stars indicate
high
statistical significance ($p<0.001$) between the male and female case.
}
\label{fig:factors-rank}
\end{figure}

Higher ranked players generally attract more media attention, and therefore may be targeted differently by journalists. To understand the effect of player ranking, we divide players into two groups: top 10 players and the rest. For our analysis, we use the ranking of the player at the time the interview was conducted.
(It is therefore possible that questions
posed to the same player
but
 at different times
could fall into different ranking groups due to ranking fluctuations over time.)
We find that questions to male players are significantly closer to game language regardless of player ranking ($p$-value $<$ 0.001, Figure \ref{fig:factors-rank}).

Furthermore, if we focus only on players who have ranked both in and outside
the
top 10 in our dataset, and pair the questions asked to them when they 
were
 higher-ranked to the questions asked when their ranking 
 was
  lower,
we
find
that there is no significant difference between questions asked to male athletes when they 
were
 in different ranking groups (Wilcoxon signed-rank $p$-value $>$ 0.05). However, the difference is significant for females (Wilcoxon signed-rank $p$-value $<$ 0.01), suggesting that gender
bias
 may be more salient for lower ranked players as questions to lower-ranked female athletes tend to 
 be less \gamerelated.

While one might 
expect that
 star players 
 would
  receive more off-court questions (yielding higher perplexities), the
 perplexity values for questions posed to top 10 players are actually lower regardless of gender. This may be because
the training data for our language model is more focused on
specific points played in matches, and may
not be representative of
tennis-related questions that are more general
(e.g., longer-term career goals, personal records,
injuries).
In other words, our result
suggests that journalists may
attend more to the specifics of the
games of higher ranked players, 
posing
 more specific questions about points played in the match during interviews.

\subsubsection{Winning vs.~Losing}
\label{sec:winning}

While it is reasonable to  
expect that %
 whether 
  the interviewee won or lost
would affect
how \gamerelated the questions are, the difference in mean perplexity
for males and females
conditioned on win/loss game outcome
are comparable. %
In addition, for both male players and female players, there is no significant difference observed between the paired set of questions asked in winning interviews and the losing ones (Wilcoxon signed-rank $p$-value $>$ 0.05), controlling for both player and season.\footnote{We pair each question asked to a given player when winning to one question posed to the {\em same} player in the {\em same} calendar year when losing to construct the paired set of winning and losing questions for each gender.} 
 This suggests
  that
 that game result may not be a factor affecting %
  how \gamerelated the interview questions are.
\section{Concluding discussion}

In this work we propose a
language-model based approach to quantify gender bias in the interview questions tennis players receive. We find that questions to male athletes are generally more
\gamerelated.
The difference is more salient among the unusual questions in press conferences, and
for lower-ranked players.

However, this preliminary study has a number of limitations.   We have
considered only a single sport.
In addition, our dataset does not contain any information about who asked which question, which makes us unable to control for any
idiosyncrasies of
specific
journalists.  For example, it is conceivable that the
disparities we observe are
explained by
differences in the journalists that are assigned to conduct the respective interviews.

In this work, we limit our scope to bias in terms of \gamerelated language, not considering differences
(or similarities)
that may exist in other dimensions. Further studies may use a similar approach to 
quantify and explore differences in other dimensions, by using language models specifically trained to model other domains of interests, which may provide a more comprehensive view of how questions differ when targeting different groups.

Furthermore, our main focus is on questions asked during press conferences;
we have not looked at
the players' responses.
The transcripts data,
which we 
 release publicly, %
may provide opportunities for further studies.
\section*{Acknowledgments} We thank the anonymous reviewers and the participants in the 
Fall
 2015 edition of the course ``Natural Language Processing and Social Interaction'' for helpful comments and discussion.
This research was supported in part by a Discovery and Innovation Research Seed award from the Office of the Vice Provost for Research at Cornell.

\bibliographystyle{named}
\bibliography{final,more-refs}

\begin{thebibliography}{}

\bibitem[\protect\citeauthoryear{Angelini and
  Billings}{2010}]{angelini2010agenda}
James~R. Angelini and Andrew~C. Billings.
\newblock An agenda that sets the frames: Gender, language, and {NBC}'s
  americanized olympic telecast.
\newblock {\em Journal of Language and Social Psychology}, 2010.

\bibitem[\protect\citeauthoryear{Billings}{2008}]{Billings:OlympicMediaInsideTheBiggestShow:2008}
Andrew~C. Billings.
\newblock {\em Olympic media: Inside the biggest show on television}.
\newblock Routledge, 2008.

\bibitem[\protect\citeauthoryear{Bruce}{2004}]{Bruce01112004}
Toni Bruce.
\newblock Marking the boundaries of the `normal' in televised sports: The
  play-by-play of race.
\newblock {\em Media, Culture \& Society}, 26(6):861--879, 2004.

\bibitem[\protect\citeauthoryear{Eastman and
  Billings}{2000}]{eastman2000sportscasting}
Susan~Tyler Eastman and Andrew~C. Billings.
\newblock Sportscasting and sports reporting: The power of gender bias.
\newblock {\em Journal of Sport \& Social Issues}, 24(2):192--213, 2000.

\bibitem[\protect\citeauthoryear{Eastman and
  Billings}{2001}]{Eastman:HowardJournalOfCommunication:2010}
Susan~Tyler Eastman and Andrew~C. Billings.
\newblock Biased voices of sports: Racial and gender stereotyping in college
  basketball announcing.
\newblock {\em Howard Journal of Communication}, 12(4):183–--201, 2001.

\bibitem[\protect\citeauthoryear{Heafield \bgroup \em et al.\egroup
  }{2013}]{kenlm}
Kenneth Heafield, Ivan Pouzyrevsky, Jonathan~H. Clark, and Philipp Koehn.
\newblock Scalable modified {Kneser-Ney} language model estimation.
\newblock In {\em Proceedings of the ACL}, pages 690--696, August 2013.

\bibitem[\protect\citeauthoryear{Higgs \bgroup \em et al.\egroup
  }{2003}]{higgs2003gender}
Catriona~T. Higgs, Karen~H. Weiller, and Scott~B Martin.
\newblock Gender bias in the 1996 olympic games: A comparative analysis.
\newblock {\em Journal of Sport \& Social Issues}, 27(1):52--64, 2003.

\bibitem[\protect\citeauthoryear{Hilliard}{1984}]{Hilliard:SociologyOfSportJournal:1984}
Dan~C. Hilliard.
\newblock Media images of male and female professional athletes: An
  interpretive analysis of magazine articles.
\newblock {\em Sociology of Sport Journal}, 1:251--262, 1984.

\bibitem[\protect\citeauthoryear{Jelinek \bgroup \em et al.\egroup
  }{1977}]{jelinek1977perplexity}
Fred Jelinek, Robert~L. Mercer, Lalit~R. Bahl, and James~K. Baker.
\newblock Perplexity --- a measure of the difficulty of speech recognition
  tasks.
\newblock {\em The Journal of the Acoustical Society of America},
  62(S1):S63--S63, 1977.

\bibitem[\protect\citeauthoryear{Jones}{2004}]{jones2004half}
Dianne Jones.
\newblock Half the story?~{Olympic} women on {ABC} news online.
\newblock {\em Media International Australia}, 110(1):132--146, 2004.

\bibitem[\protect\citeauthoryear{Kaskan and Ho}{2014}]{Kaskan:SexRoles:2014}
Emily~R. Kaskan and Ivy~K. Ho.
\newblock Microaggressions and female athletes.
\newblock {\em Sex Roles}, 74(7-8):275--287, 11 2014.

\bibitem[\protect\citeauthoryear{Kian and Clavio}{2011}]{kian2011}
Edward~M. Kian and Galen Clavio.
\newblock A comparison of online media and traditional newspaper coverage of
  the men's and women's {U.S.} open tennis tournaments.
\newblock {\em Journal of Sports Media}, 2011.

\bibitem[\protect\citeauthoryear{Kian \bgroup \em et al.\egroup
  }{2009}]{Kian:JournalOfBroadcastingElectronicMedia:2009}
Edward (Ted)~M. Kian, Mondello Michael, and Vincent John.
\newblock {ESPN} --- the women's sports network? a content analysis of internet
  coverage of {March} madness.
\newblock {\em Journal of Broadcasting \& Electronic Media}, 53(3):477--495, 9
  2009.

\bibitem[\protect\citeauthoryear{Knight and Giuliano}{2001}]{Knight:SexRoles}
Jennifer~L. Knight and Traci~A. Giuliano.
\newblock He's a {Laker}; she's a ``looker'': The consequences of
  gender-stereotypical portrayals of male and female athletes by the print
  media.
\newblock {\em Sex Roles}, 45(3-4):217--229, 2001.

\bibitem[\protect\citeauthoryear{Li{\v{c}}en and
  Billings}{2013}]{livcen2013cheering}
Simon Li{\v{c}}en and Andrew~C. Billings.
\newblock Cheering for `our' champs by watching `sexy' female throwers:
  Representation of nationality and gender in {Slovenian} 2008 {Summer}
  {Olympic} television coverage.
\newblock {\em European Journal of Communication}, 28(4):379--396, 2013.

\bibitem[\protect\citeauthoryear{Messner \bgroup \em et al.\egroup
  }{1993}]{Messner:GenderSociety:1993}
Michael~A. Messner, Margaret~Carlisle Duncan, and Kerry Jensen.
\newblock Separating the men from the girls: The gendered language of televised
  sports.
\newblock {\em Gender \& Society}, 7(1):121--137, 1993.

\bibitem[\protect\citeauthoryear{Rainville and
  McCormick}{1977}]{Rainville:JournalismMassCommunicationQuarterlyJournalism:1977}
Raymond~E. Rainville and Edward McCormick.
\newblock Extent of covert racial prejudice in pro football announcers' speech.
\newblock {\em Journalism \& Mass Communication Quarterly}, 54(1):20--26, 1977.

\bibitem[\protect\citeauthoryear{Sabo \bgroup \em et al.\egroup
  }{1996}]{sabo96}
Don Sabo, Sue~Curry Jansen, Danny Tate, Margaret~Carlisle Duncan, and Susan
  Leggett.
\newblock Televising international sport: Race, ethnicity, and nationalistic
  bias.
\newblock {\em Journal of Sport \& Social Issues}, 20(1):7--21, 1996.

\bibitem[\protect\citeauthoryear{Van~Sterkenburg \bgroup \em et al.\egroup
  }{2010}]{van2010race}
Jacco Van~Sterkenburg, Annelies Knoppers, and Sonja De~Leeuw.
\newblock Race, ethnicity, and content analysis of the sports media: {A}
  critical reflection.
\newblock {\em Media, Culture \& Society}, 32(5):819--839, 2010.

\bibitem[\protect\citeauthoryear{Vincent \bgroup \em et al.\egroup
  }{2007}]{Vincent:InternationalJournalOfSportManagementAndMarketing:2007}
John Vincent, Paul~M. Pedersen, Warren~A. Whisenant, and Dwayne Massey.
\newblock Analysing the print media coverage of professional tennis players:
  British newspaper narratives about female competitors in the {Wimbledon}
  championships.
\newblock {\em International Journal of Sport Management and Marketing},
  2(3):281--300, 2007.

\end{thebibliography}

\end{document}